%% file: misinformation_demotion.tex
\documentclass[preprint]{article}

\PassOptionsToPackage{numbers, compress}{natbib}
\usepackage{neurips_2019}

\usepackage[utf8]{inputenc} % allow utf-8 input
\usepackage[T1]{fontenc}    % use 8-bit T1 fonts
\usepackage{hyperref}       % hyperlinks
\usepackage{url}            % simple URL typesetting
\usepackage{booktabs}       % professional-quality tables
\usepackage{amsfonts}       % blackboard math symbols
\usepackage{nicefrac}       % compact symbols for 1/2, etc.
\usepackage{microtype}      % microtypography
\usepackage{booktabs} % For formal tables
\usepackage{enumitem}
\usepackage{graphicx}
\usepackage{subfig}
\usepackage{amsfonts,amssymb,amsthm,amsmath,mathrsfs,mathtools}
\usepackage{latexsym}
\usepackage{xspace}
\usepackage{Definitions}
\usepackage[utf8]{inputenc}
\usepackage[english]{babel}
\usepackage{algorithm}
\usepackage[noend]{algpseudocode}
\usepackage{bbm}
\usepackage{color}

\usepackage{url}

\usepackage{Definitions}

\title{Consequential ranking algorithms and \\ long-term welfare}

\author{%
  Behzad Tabibian \\
  MPI-IS \& MPI-SWS\\
  \texttt{me@btabibian.com} \\
  \And
  Vicen\c{c} Gomez \\
  Universitat Pompeu Fabra \\
  \texttt{vicen.gomez@upf.edu} \\
  \And
  Abir De \\
  MPI-SWS \\
  \texttt{ade@mpi-sws.org} \\
  \And
  Bernhard Sch\"{o}lkopf \\
  MPI-IS \\
  \texttt{bs@tuebingen.mpg.de} \\
  \And
  Manuel Gomez Rodriguez \\
  MPI-SWS \\
  \texttt{manuelgr@mpi-sws.org} \\
}

\DeclareMathOperator{\given}{ | }
\newcommand{\xhdr}[1]{\noindent{{\bf #1.}}}
\DeclarePairedDelimiterX{\infdivx}[2]{(}{)}{%
  #1\;\delimsize\|\;#2%
}

\newcommand{\ra}[1]{\renewcommand{\arraystretch}{#1}}
\graphicspath{{./FIG/}}

\begin{document}

\maketitle

\begin{abstract}
\input{000abstract}
\end{abstract}

\vspace{-2mm}
\section{Introduction}
\label{sec:intro}
\vspace{-1mm}
\input{010intro}

\vspace{-3mm}
\section{Rankings and User Dynamics}
\label{sec:model}
\vspace{-2mm}
\input{020model}

\vspace{-2mm}
\section{Building Consequential Rankings}
\label{sec:consequential}
\vspace{-1mm}
\input{030consequential}
\vspace{-2mm}
\section{A Stochastic Gradient-Based Algorithm}
\label{sec:algorithm}
\vspace{-1mm}
\input{040algorithm}

%\section{Experiments}
%\label{sec:experiments}
\vspace{-3mm}
\section{Experiments on synthetic data}
\label{sec:synthetic}
\vspace{-2mm}
\input{050sim}

\vspace{-3mm}
\section{Experiments on real data}
\label{sec:real}
\vspace{-2mm}
\input{060real}
\vspace{-3mm}
\section{Conclusions}
\label{sec:conclusions}
\vspace{-3mm}
\input{070conclusions}

\vspace{-2mm}

{
\small
\bibliographystyle{abbrv}
\bibliography{refs}
}

\clearpage
\newpage

\appendix
\input{080appendix}

\end{document}

%% file: 000abstract.tex
Ranking models are typically designed to provide rankings that optimize some measure of immediate utility to the
users.
As a result, they have been unable to anticipate an increasing number of undesirable long-term consequences of their
proposed rankings, from fueling the spread of misinformation and increasing polarization to degrading social discourse.
Can we design ranking models that understand the consequences of their proposed rankings and, more importantly, are
able to avoid the undesirable ones?
In this paper, we first introduce a joint representation of ran\-kings and user dynamics using Markov decision processes.
Then, we show that this representation greatly simplifies the construction of \emph{consequential ranking models} that trade off the
immediate utility and the long-term welfare.
In particular, we can obtain optimal consequential rankings just by applying weighted sampling on the rankings provided by mo\-dels that
maximize measures of immediate utility. However, in practice, such a strategy may be inefficient and impractical, specially in high dimensional
scenarios.
To overcome this, we introduce an efficient gradient-based algorithm to learn parameterized consequential ranking mo\-dels that effectively
appro\-xi\-mate optimal ones. 
We showcase our methodology using synthetic and real data gathered from Reddit and show that ranking models derived using our 
me\-tho\-do\-lo\-gy provide ranks that may mitigate the spread of misinformation and improve the civility of online discussions. 
%
% without significant deviations from the rankings provided by models maximizing measures of immediate utility.

%% file: 010intro.tex
Rankings are ubiquitous across a large variety of online services, from search engines, online shops and recommender systems to
social media and online dating.
Consequently, it has become much easier to find information, products, jobs, opinions or even potential romantic partners---rankings
have undoubtedly increased the utility users obtain from online services.
However, ranking have also been blamed to play a major role in an increasing number of missteps, particularly in the context of social
and information systems, from fueling the spread of misinformation~\cite{vosoughi2018spread}, increasing polarization~\cite{nyt2016polarization}
and degrading social discourse~\cite{guardian17} to undermining democracy~\cite{nyt17democracy}.
As the decisions taken by ranking models become more consequential to individuals and society, one must ask: what went wrong in
these cases?

% \manuel{short-sighted -> immediate. Let's be a bit diplomatic not to get people angry}
% \manuel{causal effects -> consequences. Let's avoid discussions about causality}
Current ranking models are typically designed to optimize immediate measures of utility, which often reward instant gratification.
For example, one of the guiding technical principles behind the optimization of ranking models in the information retrieval literature,
the Probability Ranking Principle (PRP)~\cite{robertson1977probability}, states that the optimal ranking should order items in terms of
probability of relevance to the user.
% (immediate) 
%
However, such measures of immediate utility do not account for long-term consequences. As a result, ranking models 
% \manuel{saving space}
% do not \emph{understand}
% the long-term consequences of their proposed rankings, which 
often have an unexpected cost to the long-term welfare.
In this work, our goal is to design consequential ranking models which understand the long-term consequences of their proposed rankings.

More specifically, we focus on a problem setting that fits a variety of real-world applications, including those mentioned previously:
at every time step, a ranking model receives a set of items and ranks these items on the basis of a measure of immediate 
utility\footnote{\scriptsize Our methodology does not need to observe the immediate utility the ranking model based their rankings on.} and a set of features.
Items may appear over time and be present at several time steps. Moreover, their corresponding features may change over time and these changes may 
be due to the influence of previous rankings.
For example, the number of likes, votes, or comments---the features---that a post---the item---published by a user receives in social media
depends largely on its ranking position~\cite{gomez2014quantifying, hodas2012visibility, kang2015vip, lerman2014leveraging}.
%
% \manuel{I removed "on the feed of the user'{}s followers" because we do not consider several feeds, just one. It fits reddit.
%
Moreover, for every sequence of rankings, there is an associated long-term (cost to the) welfare, whose specific definition is application dependent. For
example, in information integrity, the welfare may be defined as the average number of posts including misinformation at the top of the rankings over time.
Then, our goal is to construct consequential ranking models that optimally trade off fidelity to the original ranking model maximizing immediate utility and
long-term welfare.

\xhdr{Our contributions}
In this paper, we first introduce a joint representation of ranking models and user dynamics using Markov decisions processes (MDPs), which is particularly well-fitted
to faithfully characterize
the above problem setting\footnote{\scriptsize In this work, for ease of exposition, we assume all users get exposed to the same rankings, as in, \eg, Reddit. However, our methodology
can be readily extended to the scenario in which each user get exposed to a different ranking, as in, \eg, Twitter.}.
Then, we show that this representation greatly simplifies the construction of consequential ranking models that trade off fidelity to the rankings provided by a model
maximizing immediate utility and the long-term welfare.
More specifically, we apply Bellman'{}s principle of optimality and show that it is possible to derive an analytical expression for the optimal consequential ranking model in terms
of the original ranking model and the cost to the welfare.
This means that we can obtain optimal consequential rankings just by applying weighted sampling on the rankings provided by the original ranking
model using the (exponentiated) cost to welfare.
However, in practice, such a naive sampling will be inefficient, specially in the presence of high-dimensional features.
Therefore, we design a practical and efficient gradient-based algorithm to learn parameterized consequential ranking models that effectively approximate optimal ones\footnote{\scriptsize We will release an open-source implementation of our algorithm with the final version of the paper.}.

Finally, we showcase our methodology using synthetic and real data gathered from Reddit. Our results show that consequential ranking models derived using our 
methodology provide ranks that may mitigate the spread of misinformation and improve the civility of online discussions without significant deviations from the rankings 
provided by models maximizing immediate utility measures.

\xhdr{Further related work}
%
% \manuel{saving space}
% Our work relates to several lines of research: (i) ranking algorithms; (ii) delayed impact of machine learning algorithms; (iii) optimal control and reinforcement learning;
% and, (iv) reducing the spread of misinformation and polarization.
%
% \noindent --- \emph{Ranking algorithms:} 
%
The work most closely related to ours is devoted to construct either fair rankings~\cite{asudehy2017designing, biega2018equity, celis2017ranking, singh2017equality, singh2018fairness, singh2019policy, yang2017measuring, zehlike2017fa} or diverse rankings~\cite{carbonell1998use, clarke2008novelty, radlinski2009redundancy, radlinski2008learning}.
However, this line of research defines fairness and diversity in terms of exposure allocation on an individual ranking.
In contrast, we consider sequences of rankings, we characterize the consequences of these rankings on the user dynamics, and focus on improving the welfare in the long-term.
%
% VICENC: replaced this: Our work also relates to a recent line of work that looks at the design of ranking algorithms from the perspective of reinforcement learning~\cite{feng2018greedy, singh2019policy, wei2017reinforcement}. 
%
Other related work includes recent approaches to address the learning-to-rank problem from the perspective of reinforcement learning~\cite{singh2019policy, feng2018greedy, wei2017reinforcement}. 
%
% VICENC : replaced this: However, these approaches are orthogonal to ours since they model the construction of a single ranking as an MDP, where each time step corresponds to a ranking position. In contrast, we model the construction of a sequence of rankings as an MDP.
However, these approaches consider the construction of a single optimal ranking as an MDP in which the state defined at the level of an item/position.
In contrast, our MDP considers a sequence of rankings.
%\cite{}

%
% \noindent --- \emph{Delayed impact of ML algorithms:} the 
%
Finally, there is a paucity of work on the delayed impact of machine learning algorithms~\cite{hucheng2018,liu18delayed,mouzannar2019fair} and recommender systems~\cite{sinha2016deconvolving,schnabel2018short}.
However, the former has focused on classification tasks and considered simple one-step feedback models and the latter on the tradeoff between exploration and exploitation. 

% \noindent --- \emph{Optimal control and reinforcement learning:} the work most closely related to ours within the extensive literature on optimal control and reinforcement learning
% is devoted to improving the functioning of social and information systems~\cite{behzad2019pnas, upadhyay2018deep, wang2017variational, control18jmlr, redqueen17wsdm, thalmeier2016action}. However, 
% this line of work has mainly focused on representations based on temporal point processes and have not considered rankings. 
% Recently, a framework based on survival process has been proposed to optimize click through rate using reinforcement learning~\cite{Zheng2018}.

% \manuel{Saving space}
% \noindent --- \emph{Reducing the spread of misinformation and polarization:} the literature on algorithms for reducing the spread of misinformation~\cite{balmau2018limiting, kim2018leveraging, 
% tschiatschek2018fake, vo2018rise} and reducing polarization~\cite{garimella2017reducing, garimella2017balancing, babaei2018purple, lex2018mitigating, musco2018minimizing}
% is expanding very rapidly (refer to Kumar and Shah~\cite{kumar2018false} for an excellent review of recent work).
%
% However, to the best of our knowledge, previous work has not approached the problem from the perspective of ranking algorithms.

%% file: 020model.tex
In this section, we first introduce our joint representation of rankings and user dynamics, starting from the problem setting
it is designed for.
Then, we formally define consequential rankings as the solution to a particular reinforcement learning problem.

\xhdr{Problem setting}
Let $p_{\theta}$ be a particular ranking model\footnote{\scriptsize Unless stated otherwise, the notation $p_{\theta}$ does not imply that $\theta$ is a
parameter within a class of ranking models, but it just serves as a placeholder to identify a specific ranking model.} (or, equivalently,
ranking algorithm).
At each time step $t \in \{1, \ldots, T\}$, the ranking model receives a set of $n$ items and these items are
characterized by a feature matrix $\Xb(t) \in \RR^{n \times p}$, where the $i$-th row $\Xb_i(t)$ contains
the feature values for item $i \in [n]$ and $p$ is the number of features per item.
Here, we assume that items may appear over time and be present at several time steps. Moreover, their corresponding feature values may change over time.
%
% Here, we assume that, while some of the items may be present at se\-ve\-ral time steps, their corresponding
% features may change over time.
%
For example, think of the number of likes, votes or comments that a post receives in social media---they are
often used as features to decide the ranking of the post and they change over time.

Then, the ranking model provides a ranking $\yb(t)$ of the items on the basis of their set of features and a (hidden) measure
of immediate utility. A ranking $\yb(t) = (y_1(t), \ldots, y_n(t))$ is defined as a permutation of the $n$ rank indices, \ie,
the model ranks item $i$ in position $y_i(t)$, where highest rank is position $1$.
In addition, we also define the ordering $\omegab(t) = (\omega_1(t), \ldots, \omega_n(t))$ of a ranking as a permutation
of the $n$ item indices, \ie, the model ranks item $\omegab_i(t)$ in position $i$.
The ranking and orderings are related by $w_{y_i(t)}(t) = i$ and $y_{w_i(t)}(t) = i$.
Here, we assume that the provided ranking at time step $t$ may influence the feature matrix at time step $t+1$. This is
in agreement with recent empirical studies~\cite{gomez2014quantifying, hodas2012visibility, kang2015vip, lerman2014leveraging},
which have shown that the posts (the items) that are ranked highly receive a higher number of likes, comments or shares (the features).

Finally, given a trajectory of feature matrices and rankings $\tau = \{ (\Xb(t), \yb(t)) \}_{t=0}^{T}$ there is an additive cost to the welfare,
$c(\tau) = \sum_{t=0}^{T} c(\Xb(t), \yb(t))$, where $c(\Xb(t), \yb(t))$ is an arbitrary immediate cost whose specific definition is application
dependent.
For example, in information integrity, the welfare may be defined as the average number of posts including misinformation at the top of the
rankings over time.
In the remainder,
%
% \manuel{we never used that notation below}
% whenever needed, we will use $\tau(\theta)$ to make the dependence on the ranking model $\theta$ used in the trajectory explicit and
%
we will say that a trajectory $\tau$ is \emph{induced} by a ranking model $p_{\theta}$.

%
% In the remainder, whenever needed, we will use $\tau(\theta)$ to make the dependence on the ranking model $\theta$ used in the trajectory explicit.
%

\xhdr{Joint representation of rankings and user dynamics}
The above problem setting naturally fits the following joint representation of rankings and user dynamics using Markov decision processes (MDPs)~\cite{sutton2018reinforcement},
which also has an intuitive causal interpretation:
%
% \manuel{saving space} \prod_{t=1}^{T} p_{\theta}(\Xb(t), \yb(t) \given \Xb(t-1), \yb(t-1)) =
\begin{equation} \label{eq:model}
p_{\theta}(\tau \given \Xb(t_0), \yb(t_0)) = \prod_{t=1}^{T} \underbrace{p_\theta(\yb(t) \given \Xb(t))}_{\text{ranking model}} \, \underbrace{p(\Xb(t) \given \Xb(t-1), \yb(t-1))}_{\text{user dynamics}},
\end{equation}
where the first term represents the particular choice of ranking model\footnote{\scriptsize In our work, we consider probabilistic ranking models, which assign
a probability to each ranking. It would be interesting to extend our methodology to deterministic ranking models.},
%
% \manuel{I remove that it is parameterized by \theta, since here I do not want to tell p_{\theta} is a parametric family since otherwise, the optimal
% consequential ranking model is implicitly assumed to be of a particular family.}
% parameterized by $\theta$,
%
the second term represents the distribution for the user dynamics, which determines the feature matrix at any given time step, and the initial feature matrix
$\Xb(t_0)$ and ranking $\yb(t_0)$ are given. % given the previous feature matrix and ranking.
Moreover, the above representation makes two major assumptions, which are also illustrated in Figure~\ref{fig:representation} in
Appendix~\ref{app:model}: %, which we believe hold in most practical scenarios:
\begin{itemize}[noitemsep,nolistsep,leftmargin=0.8cm]
\item[(i)] To provide a ranking for a set of items at time step $t$, the ranking model only uses the feature matrix corresponding to that set of items.
More formally, given the feature matrix $\Xb(t)$, the ranking $\yb(t)$ provided by the ranking model is conditionally independent of previous feature
matrices $\Xb(t')$, $t' < t$. In most practical scenarios, ranking models optimizing for immediate utility satisfy this assumption.
\item[(ii)] The dynamics of the feature matrices, which characterize the user dynamics, are Markovian. That means, given the feature matrix $\Xb(t-1)$
and ranking $\yb(t-1)$, the feature matrix $\Xb(t)$ is conditionally independent of previous feature matrices $\Xb(t')$ and rankings $\yb(t')$, $t' < t$.
%
%Most state of the art models of users dynamics (\eg, Hawkes processes with exponential kernels) do assume the Markov property~\cite{de2016learning, du2016recurrent, behzad2019pnas}.
This is a natural assumption taken in %most of
 the state-of-the-art models (\eg, Hawkes processes with exponential kernels~\cite{de2016learning, du2016recurrent, behzad2019pnas}).

%
% , however, in practice, this assumption may be violated for some choice of features.
%
\end{itemize}
%
% It would be very interesting, albeit challenging, to lift the second assumption in future work.
%
% \manuel{saving space}
% Next, we elaborate further on the specifics of the ranking model and the distribution of the user dynamics.

\vspace{1mm} \noindent \emph{--- Ranking model:} Our approach is agnostic to the particular choice of ranking model---it provides a methodology to derive consequential
rankings that are optimal under a ranking model.
In our experiments, we showcase our methodology for one well-known ranking model, the Plackett-Luce (P-L) ranking model~\cite{luce1977choice, plackett1975analysis},
which is best described in terms of the orderings of the rankings. More specifically, under the P-L model, at each time step $t$, the ranking $\yb(t)$, with ordering $\omegab(t)$, is
sampled from a distribution
\begin{equation} \label{eq:pl}
p_{\thetab}(\yb(t) \given \Xb(t)) = \prod_{k=1}^{n} f_k(\Xb(t)) = \prod_{k=1}^{n} \frac{\exp\left(\thetab^{T} \Xb_{\omega_k}(t)\right)}{\sum_{k'=k}^{N} \exp\left(\thetab^{T} \Xb_{\omega_{k'}}(t)\right)}
\end{equation}
%
%where
%
%\begin{equation} \label{eq:fk}
%f_k(\Xb(t)) =
%\end{equation}
%
where $\thetab$ is a given parameter.
In the above, we can think of $\thetab^{T} \Xb_{\omega_k}(t)$ as a \emph{quality score} associated to the item $\omega_k$, which controls the probability
that this item is ranked at the top---the higher the quality score, the higher the probability that the item is ranked first.
In practice, the quality score of the above P-L ranking model may be computed using a complex nonlinear function~\cite{tran2016choice}, \eg, a neural
network.
%
% \manuel{This is for the experiments. It is confusing to mention this here since, in practice, we only need that the ranking model gives us
% the value of p_{\thetab}(\yb(t) \given \Xb(t)) for any \Xb(t). We do not necessarily need to _fit_ a ranking model to historical ranking data}
% Moreover, in practice, the parameter $\wb$ can be estimated from historical rankings by the ranking model using maximum likelihood
% estimation (MLE)~\cite{hunter2004mm} or approximate deterministic Bayesian inference~\cite{minka2005divergence}.

\vspace{1mm} \noindent \emph{--- User dynamics:} Our approach only requires to be able to sample $\Xb(t)$ from any arbitrary model for the
transition probability $p(\Xb(t) \given \Xb(t-1), \yb(t-1))$, which may be estimated using historical ranking and user data.
Here, in contrast with the ranking model, the user dynamics are not something that one can decide upon---they are given.

%; in other words, we will only need to be able to sample from $p(\Xb(t) \given \Xb(t-1), \yb(t-1))$.

\xhdr{Consequential rankings}
Let $p_{\theta_0}$ be an existing ranking model that optimizes some hidden immediate utility and $c(\cdot)$ a cost to the welfare.
Then, we construct a consequential ranking model $p_{\theta}$, which optimally trades off
the fidelity to the original ranking model and the cost to the long-term welfare, by solving the following optimization problem:
\begin{align} \label{eq:optimization-problem}
\underset{p_{\theta}}{\text{minimize}} & \quad \EE_{\tau \sim p_{\theta}} \left[S_{\theta}(\tau \given \Xb(0), \yb(0))\right],
\end{align}
with
\begin{equation} \label{eq:S}
S_{\theta}(\tau \given \Xb(0), \yb(0)) = c(\tau) + \lambda \log \frac{p_{\theta}(\tau \given \Xb(0), \yb(0))}{p_{\theta_0}(\tau \given \Xb(0), \yb(0))},
\end{equation}
where the expectation is taken over all the trajectories $\tau = \{ (\Xb(t), \yb(t)) \}_{t=0}^{T}$ of feature matrices and rankings of length $T$ under the ranking
model $p_{\theta}$ and $\lambda \geq 0$ is a given parameter which controls the trade off between the fidelity to the original ranking model and the long-term
cost to the welfare. % , and we do not assume any specific parametric form for the consequential ranking model $p_{\theta}$.
In Eq.~\ref{eq:S}, the first term penalizes trajectories that achieve a large cost to the welfare and the second term penalizes ranking models whose induced
trajectories differ more from those that the original model would induce, where the terms associated to the user dynamics $p(\Xb(t) \given \Xb(t-1), \yb(t-1))$
cancel.
Moreover, the choice of trajectory length $T$ will depend on the definition of long-term---accounting for longer-term consequences to the welfare will require
larger trajectory lengths $T$. % \footnote{\scriptsize Our methodology can be readily augmented to trajectories of infinite length (or horizon) with discounted costs.}.

Finally, note that our measure of fidelity has a natural interpretation in terms of the Kullback-Leibler (KL) divergence~\cite{kullback1951information}, which has
been extensively used as a \emph{distance} measure between probability  distributions, leading to the formulation of reinforcement learning as probabilistic inference~\cite{levine2018reinforcement,kappen2012optimal,ziebartICML}.
%distributions in the reinforcement learning literature~\cite{kappen2012optimal,levine2018reinforcement,ziebartICML}.
%
More specifically, we can write the expectation of the second term as the KL divergence between the original and the consequential ranking model,
\ie, $KL[p_{\theta}(\cdot \given \Xb(0), \yb(0)) \, || \, p_{\theta_0}(\cdot\given \Xb(0), \yb(0))]$.
 % = \EE_{\tau \sim p_{\theta}} \left[\log \frac{p_{\theta}(\tau \given \Xb(0), \yb(0))}{p_{\theta_0}(\tau \given \Xb(0),
% \yb(0))}\right]$.
%
% \manuel{If there is space, uncomment next line}
% In the next section, we will exploit this interpretation to greatly simplify the construction of consequential rankings.

%% file: 030consequential.tex
In this section, we tackle the optimization problem defined by Eq.~\ref{eq:optimization-problem} from the perspective of reinforcement
learning and show that the optimal consequential ranking model can be expressed in terms of the original ranking model.

We can first break the above problem into small recursive subproblems using Bellman'{}s principle of optimality~\cite{bertsekas}. This readily follows
from the fact that, under the representation introduced in Section~\ref{sec:model}, the ranking model and the user dynamics are a Markov decision
process (MDP).
More specifically, Bellman'{}s principle tells us that the optimal ranking model should satisfy the following recursive equation, which is called the
Bellman optimality equation:
\begin{equation}
V_t(\Xb, \yb) = \min_{p_{\theta}} \, \ell(\Xb, \yb) + \EE_{(\Xb', \yb') \sim p_{\theta}(\Xb', \yb' \given \Xb, \yb)} \left[V_{t+1}(\Xb'{}, \yb'{})\right] \label{eq:bellman}
\end{equation}
with $V_T(\Xb, \yb) = \ell(\Xb, \yb)$. The function $V_t(\Xb, \yb)$ is called the value function and the function $\ell(\Xb, \yb)$ is called immediate loss. Moreover,
in our problem, it can be readily shown that the immediate loss adopts the following form:
\begin{align*}
\ell(\Xb, \yb) &= c(\Xb, \yb) + \lambda \EE_{(\Xb'{}, \yb'{}) \sim p_{\theta}(\Xb', \yb' \given \Xb, \yb)} \left[\log \frac{p_{\theta}(\Xb', \yb' \given \Xb, \yb)}{p_{\theta_0}(\Xb', \yb' \given \Xb, \yb)}\right] \\
&= c(\Xb, \yb) + \lambda KL( p_{\theta}(\cdot\,,\, \cdot \given \Xb, \yb) \, || \, p_{\theta_0}(\cdot\,,\, \cdot \given \Xb, \yb)).
\end{align*}
Within the loss function, the first term penalizes the immediate cost to the welfare and the second term penalizes consequential ranking models whose induced transition probability
differs from that induced by the original ranking model.

In general, Bellman optimality equations are difficult to solve. However, the structure of our problem will help us find an analytical solution.
Inspired by Todorov~\cite{todorov2009efficient}, we proceed as follows.
Let $Z_t(\Xb, \yb) = \exp(-V_t(\Xb, \yb))$. Then, we can rewrite the minimization in the RHS of Eq.~\ref{eq:bellman} as
\begin{equation*}
\min_{p_\theta} \EE_{(\Xb', \yb') \sim p_{\theta}(\Xb', \yb' \given \Xb, \yb)} \left[ \log \frac{p_{\theta}(\Xb', \yb' \given \Xb, \yb)}{p_{\theta_0}(\Xb', \yb' \given \Xb, \yb) Z_{t+1}(\Xb', \yb')} \right],
\end{equation*}
where we have dropped $\lambda$ and $c(\Xb, \yb)$ because they do not depend on $p_\theta$.
Then, we can use Eq.~\ref{eq:model} to factorize both transition probabilities in the numerator and the denominator within the logarithm
and, as a result, the terms $p(\Xb' \given \Xb, \yb)$ cancel and we obtain:
\begin{equation*}
\min_{p_\theta} \EE_{(\Xb', \yb') \sim p_{\theta}(\Xb', \yb' \given \Xb, \yb)} \left[ \log \frac{p_{\theta}( \yb' \given \Xb')}{p_{\theta_0}(\yb' \given \Xb') Z_{t+1}(\Xb', \yb')} \right].
\end{equation*}
The above equation resembles a KL divergence, however, note that the fraction within the logarithm does not depend on $(\Xb, \yb)$ and the
denominator $p_{\theta_0}(\yb' \given \Xb') Z_{t+1}(\Xb', \yb')$ is not normalized to one.
If we multiply and divide the fraction by the following normalization term:
\begin{equation}
G[Z_{t+1}] (\Xb') = \EE_{ \yb' \sim p_{\theta_0}(\yb' | \Xb')} [ Z_{t+1}(\Xb', \yb') ],
\end{equation}
we obtain:
\begin{align*}
\min_{p_\theta} &- \EE_{\Xb' \sim p(\Xb' \given \Xb, \yb)} \left[ \log G[Z_{t+1}](\Xb') \right]
+ \EE_{(\Xb', \yb') \sim p_{\theta}(\Xb', \yb' \given \Xb, \yb)} \left[ \log \frac{p_{\theta}( \yb' \given \Xb') G[Z_{t+1}](\Xb')}{p_{\theta_0}(\yb' \given \Xb') Z_{t+1}(\Xb', \yb')} \right].
%&+ KL\left( p_{\theta}(\cdot\,,\, \cdot \given \Xb, \yb) \,||\, \frac{p_{\theta_0}(\cdot\,,\, \cdot \given \Xb, \yb) Z_{t+1}(\cdot\,,\, \cdot)}{G[Z_{t+1}](\Xb, \yb)} \right)
\end{align*}

In the above equation, note that the first term does not depend on $p_{\theta}$ and the second term achieves its global minimum of zero if the numerator
and the denominator are equal. Thus, the optimal consequential ranking model is just given by:
\begin{equation} \label{eq:optimal-consequential-ranking-model}
p^{*}_\theta(\yb \given \Xb) = \frac{p_{\theta_0}(\yb \given \Xb) Z_{t+1}(\Xb, \yb)}{G[Z_{t+1}](\Xb)}.
\end{equation}
Finally, if we substitute back $p^{*}_{\theta}$ into the Bellman equation, given by Eq.~\ref{eq:bellman}, and write it in terms of $Z_t$,
we can also find the function $Z_t$ using the following recursive expression:
\begin{equation*}
Z_t(\Xb, \yb) = \exp \left( -c(\Xb, \yb) + \lambda \EE_{\Xb' \sim p(\Xb' \given \Xb, \yb)} \left[ \log G[Z_{t+1}](\Xb') \right] \right),
\end{equation*}
with $Z_T(\Xb, \yb) = -\log c(\Xb, \yb)$.
The above result has an important implication.
It means that we can use sampling methods to obtain (unbiased) samples from the optimal consequential ranking, \eg,
stratified sampling~\cite{douc2005comparison}, as shown in Appendix~\ref{app:sampling-alg}.
However, in practice, these sampling methods may be inefficient and have high variance if the original ranking model $p_{\theta_0}$
produces rankings that have very low probability under the optimal consequential ranking model. This will be specially problematic in
the presence of high-dimensional feature vectors due to the curse of dimen\-sio\-nality.
In the next section, we will present a practical method for approximating $p^{*}_\theta(\yb \given \Xb)$, which iteratively adapts a parameterized
consequential ranking model using a stochastic gradient-based algorithm.

%% file: 040algorithm.tex
%
%%%%%%%%%%%%%%%%%%%%%%%%%%%%%%%%%%%%%%%%%%%%%%%%%%%%%%%%%%%%%%%%%%%%%%%%%%%%%%
\begin{algorithm}[t!]
\small
\caption{It trains a parameterized consequential ranking model.}
\label{alg:consequential-ranking}
  \begin{algorithmic}[1]
    \Require Cost to welfare $c(\cdot)$, parameter $\lambda$, original ranking model $p_{\theta_0}$, $(\Xb(0), \yb(0))$, \# of iterations $M$, mini batch size $B$, and learning rate $\gamma$.
%    stationary true distributions $P(x,s)$ and conditional $P(y \given x,s)$; fixed feature map $\phi$;
   \Statex
%     \Function{TrainPolicies}{$c$, $T$, $N$, $M$, $B$, $\gamma$}
        \State $\theta^{(0)} \gets \Call{InitializeRankingModel}{{}}$
        \For{$j=1,\ldots, M$} \Comment{iterations}
        \State $\Dcal \gets \Call{Minibatch}{{p_{\theta}, B}}$ \Comment{sample mini batch}
        \State $\nabla \gets 0$
        \For{$\tau^{(i)} \in \Dcal$}
                 \State $S \gets c(\tau^{(i)}) + \lambda \log \frac{p_{\theta^{(j)}}(\tau^{(i)} \given \Xb(0), \yb(0))}{p_{\theta_0}(\tau^{(i)} \given \Xb(0), \yb(0))}$
                \State $\nabla \gets \nabla + \left( S + \lambda \right) \nabla_\theta \log p_{\theta^{(j)}}(\tau^{(i)} \given \Xb(0), \yb(0) )$
         \EndFor
         \State $\theta^{(j+1)} \gets \theta^{(j)} + \gamma \, \frac{\nabla}{B}$
        \EndFor
        \State \Return $\theta^{(M)}$
  \end{algorithmic}
\end{algorithm}
In this section, our goal is to find a consequential ranking model $p_{\theta}$ within a class of parameterized ranking models $\Pcal(\Theta)$
that approximates well the optimal consequential ranking model $p^{*}_{\theta}$, given by Eq.~\ref{eq:optimal-consequential-ranking-model},
that minimizes the objective function in Eq.~\ref{eq:optimization-problem}, \ie, $\EE_{\tau \sim p_{\theta}} \left[S_{\theta}(\tau \given \Xb(0), \yb(0))\right]$.

To this aim, we introduce a general gradient-based algorithm, which only requires the class of parameterized ranking models $\Pcal(\Theta)$ to be differentiable.
In particular, we resort to stochastic gradient descent (SGD)~\citep{kiefer1952stochastic}, \ie, $\theta^{(j + 1)} = \theta^{(j)} + \gamma_{j} \nabla_{\theta}\left. \EE_{\tau \sim p_{\theta}} \left[S_{\theta}(\tau \given \Xb(0), \yb(0))\right] \right|_{\theta = \theta^{(j)}}$, where $\gamma_j > 0$ is the learning rate at step $j \in \NN$.
%
% \footnote{\scriptsize Depending on the choice of the class of parameterized ranking models and the learning rate schedule, our algorithm may enjoy theoretical guarantees,
% which we leave for future work. Here, we will demonstrate that our algorithm does perform well in practice.}
%
Here, it may seem challenging to compute a finite sample estimate of the gradient of the objective function $\EE_{\tau \sim p_{\theta}} \left[S_{\theta}(\tau \given \Xb(0), \yb(0))\right]$
since the derivative is taken with respect to the parameters of the ranking model $p_{\theta}$, which we are trying to learn.
However, we can overcome this challenge using the log-derivative trick as in~\citep{williams1992simple}, which allows us to write the gradient as:
\begin{align} \label{eq:gradient}
\nabla_{\theta} & \EE_{\tau \sim p_{\theta}} \left[ S_{\theta}(\tau \given \Xb(0), \yb(0)) \right] = \EE_{\tau \sim p_\theta}\left[ \left( S_\theta(\tau \given \Xb(0), \yb(0)) + \lambda \right) \nabla_\theta \log p_\theta (\tau \given \Xb(0), \yb(0) ) \right],
\end{align}
where $\nabla_\theta \log p_\theta (\tau \given \Xb(0), \yb(0) )$ is often referred as the score function~\citep{Hyvarinen05:ScoreMatching}.
%
% \manuel{saving space}
% This yields the following unbiased finite sample Monte-carlo estimator for the gradient:
%
% \begin{equation} \label{eq:empirical-gradient}
% \sum_{i=1}^{B} \left( S_\theta(\tau^{(i)} \given \Xb(0), \yb(0)) + \lambda \right) \nabla_\theta \log p_\theta (\tau^{(i)} \given \Xb(0), \yb(0) ),
% \end{equation}
%
% where $B$ is the number of sampled trajectories from the joint distribution $p_{\theta}(\tau \given \Xb(0), \yb(0))$ induced by the ranking model $p_{\theta}$.
%
The overall procedure is summarized in Algorithm~\ref{alg:consequential-ranking}, where \textsc{Minibatch}$(p_{\theta}, B)$ samples a minibatch of size $B$ from $p_{\theta}(\tau)$ and \textsc{InitializeRankingModel}$()$ initializes the parameters of the ranking model.

\xhdr{Remarks} Note that, to compute an empirical estimate of the gradient in Eq.~\ref{eq:gradient}, we only need to be able to sample from the user dynamics $p(\Xb(t) \given \Xb(t-1), \yb(t-1))$,
since the explicit dependence cancels out within $S_{\theta}(\tau \given \Xb(0), \yb(0))$, as pointed out in Section~\ref{sec:model}. Moreover, depending on the choice of parameterized family of
ran\-king models, one may be able to compute the score functions analytically.
In our experiments, the class of Plackett-Luce (P-L) ranking models allows for that, \ie, % . More specifically, it readily follows from Eq.~\ref{eq:pl} that
%
%\nabla_\theta \sum_{t=1}^{T}  \sum_{k=1}^{n} \log f_k(\Xb(t))
\vspace{-1mm}
\begin{equation*}
    \nabla_\theta \log p_\theta (\tau \given \Xb(0), \yb(0) ) % &=\nabla_\theta \sum_{k=1}^{n}  \bigg(\theta^T \Xb_{\omega_{k}}(t) - \log \sum_{k'=k}^{n} \exp(\theta^T \Xb{\omega_{k'}}(t)) \bigg)\\
= \sum_{t=1}^{T}\sum_{k=1}^{n} \bigg(\theta^T - \nabla_\theta \log\sum_{k'=k}^{n} \exp(\theta^T \Xb{\omega_{k'}}(t))\bigg),
\end{equation*}
where the second term within the logarithm in the last equation is the derivative of the log-sum-exp function, whose analytical expression can be
found elsewhere.
Finally, if we think of the para\-meterized ranking model $p_{\theta}$ as a policy, our algorithm resembles policy gradient algorithms used in the reinforcement learning
literature~\cite{sutton2018reinforcement}.
This connection opens up the possibility of using variance reduction techniques used in policy gradient to improve the empirical estimation of the gradient~\cite{zhao2011analysis}.
%
% This is left as future work.

% Finally, for a fixed trajectory $\tau$ the cost function $S_{\theta}(\tau \given \Xb(0), \yb(0))$ can be computed without requiring access to the model for user dynamics $p(\Xb(t+1) \given \Xb(t), \yb(t))$ since the
% terms associated with user dynamics cancel in Eq.~\ref{eq:optimization-problem} and we obtain $S_{\theta}(\tau \given \Xb(0), \yb(0))$ as:
% \begin{equation*}
% \EE_{\tau \sim p_{\theta}} \left[S_{\theta}(\tau \given \Xb(0), \yb(0))\right]  = \EE_{\tau \sim p_{\theta}} \left[ \log \frac{\prod_{t=1}^n p_{\theta}( \yb(t) \given \Xb(t))}{\prod_{t=1}^n p_{\theta_0}( \yb(t) \given \Xb(t))}
% \right].
% \end{equation*}
%
% This means that we can train our model as long as we only have access to samples from user dynamics $p(\Xb(t+1) \given \Xb(t), \yb(t))$.

%% file: 050sim.tex
In this section, our goal is compare the performance of ranking models that maximize an immediate measure of utility against consequential rankings
% \manuel{saving space}
% models, given by Eq.~\ref{eq:optimal-consequential-ranking-model}, and parameterized consequential rankings models, learned using Algorithm~\ref{alg:consequential-ranking},
in a problem setting with known user dynamics satisfying the Markov property.

\xhdr{Experimental setup}
Each trajectory has length $T = 30$ and, at each time step $t \in \{0, \ldots, T\}$, the ranking model receives a set of
$n=10$ posts $\Ical(t)$ and ranks them.
Given a set of items $\Ical(t)$ and a ranking $\yb(t)$, we assume that the set of items $\Ical(t+1)$ is just a copy of the
set of items $\Ical(t)$ where the $d \sim \text{Poisson}(1)$ posts at the bottom of the ranking $\yb(t)$ are replaced by
new posts.
Each post $i$ has two features $\Xb_i(t) = [p_i, a_i(t)]$, where $p_i$ is the (static) probability that the post
is misinformation and $a_i(t)$ is the (dynamic) rate of shares at time $t$, initialized with  $a_i(0)=0$.
There are high risk posts ($p_i = 0.6$) and low risk posts ($p_i = 0.1$) and a post is either high risk or low risk uniformly
at random.
Thus, whether the actual post is misinformation or not is a latent variable $m_i \sim \text{Bernouilli}(p_i)$, which is unobserved
by the ranking model.
The instantaneous rate of shares for each item $i$ is given by
%
%\begin{align}\label{eq:dyn}
$a_{i}(t+1)= \exp(-2 (t-s_i)) \left(a_{i}(t)+ \alpha_i + 0.1 y_i(t)\right)$,
%\end{align}
%
where $s_i$ is the time when the post was first ranked by the ranking model, $\alpha_i$ is the virality, and a
post is either viral ($\alpha_i = 10$) or non viral ($\alpha_i = 0.1$) uniformly at random.
Here, note that rate of shares of an item increases if the item is ranked at the top, as observed in previous
empirical studies.

The original ranking model $p_{\theta_0}$ aims to promote viral posts on the top $K = 3$ positions of the ranking at each time
$t$, \ie, % its immediate utility $u(t)$ is defined as
%
% \begin{equation} \label{eq:synthetic-u}
$u(t) = \sum_{k=1}^{K=3} a_{\omega_k(t)}(t)$,
% \end{equation}
%
where $\omegab(t)$ is the ordering of the ranking $\yb(t)$.
To this aim, it uses a Plackett-Luce (P-L) model, given by Eq.~\ref{eq:pl}, with $\theta \in [0, 1]$.
The consequential ranking models $p_{\theta}$ aim to trade off fidelity to the original model and the long-term presence of misinformation
on the top $K = 3$ positions of the rankings, \ie, % its cost to welfare is defined as
%
% \begin{equation} \label{eq:synthetic-c}
$c(\tau) = \frac{1}{T} \sum_{t=1}^{T} \sum_{k=1}^{K=3} p_{\omega_k(t)}$.
% \end{equation}
%
Here, we consider two consequential rankings models: (i) an optimal consequential ranking model~$p^{*}_{\theta}$, which provides rankings by applying weighted sampling
on the rankings provided by the original ranking model $p_{\theta_0}$;
and, (ii) a Plackett-Luce (P-L) consequential ranking model $p_{\theta}$, which is learned using Algorithm~\ref{alg:consequential-ranking} with $M = 400$ iterations and $B = 50$ as batch size.
Moreover, we experiment with different values of the parameter $\lambda$, which controls the trade off between fidelity to the original model
and cost to the welfare and, for each experiment, we perform $200$ repetitions.
\begin{figure*}[t]
\setlength{\tabcolsep}{0.05em}
 \centering
 \subfloat[Immediate utility vs time]{\includegraphics[width = 0.25\textwidth]{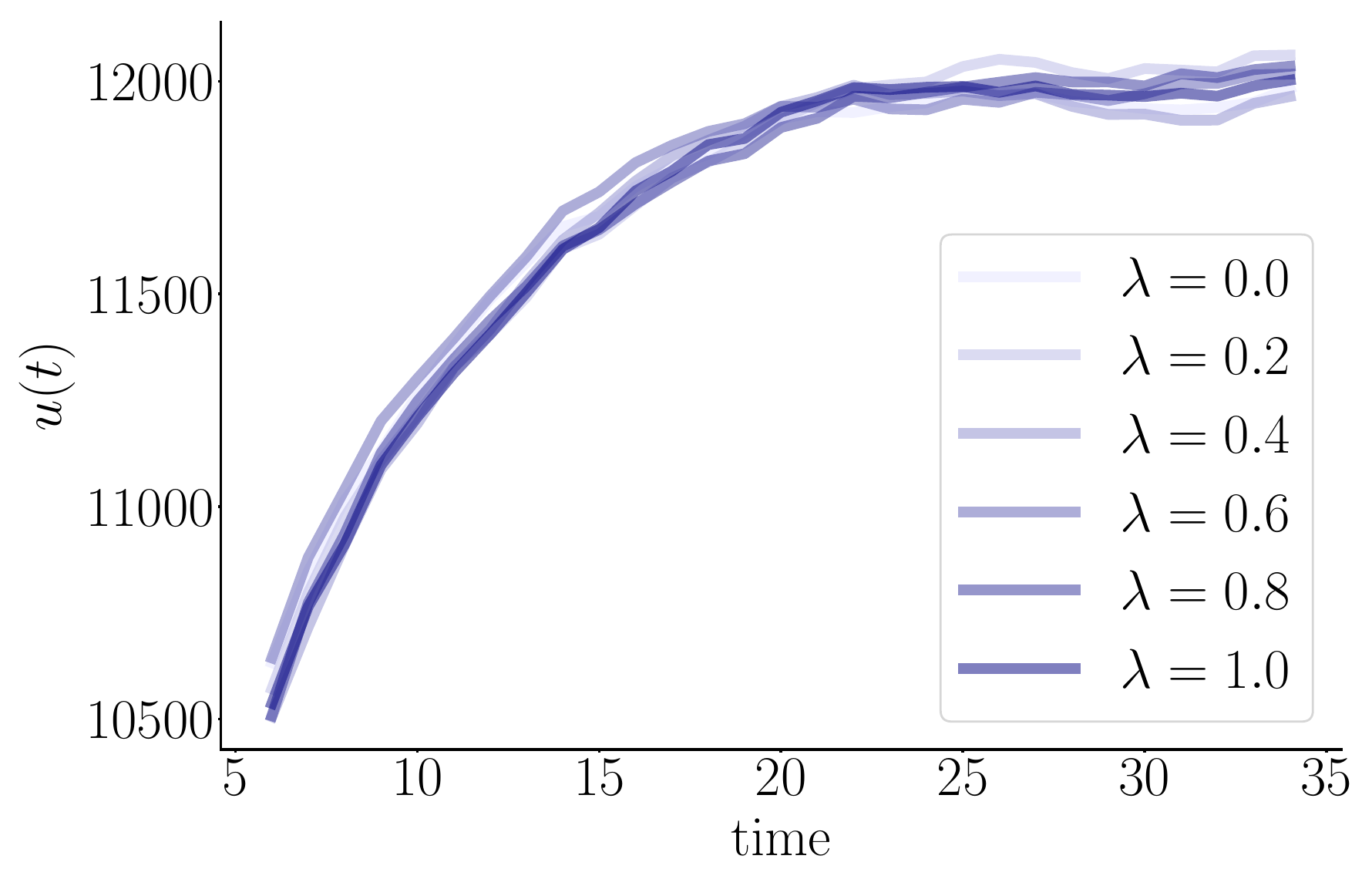}} \hspace{6mm}
 \subfloat[Cost to welfare vs $1/\lambda$]{\includegraphics[width = 0.25\textwidth]{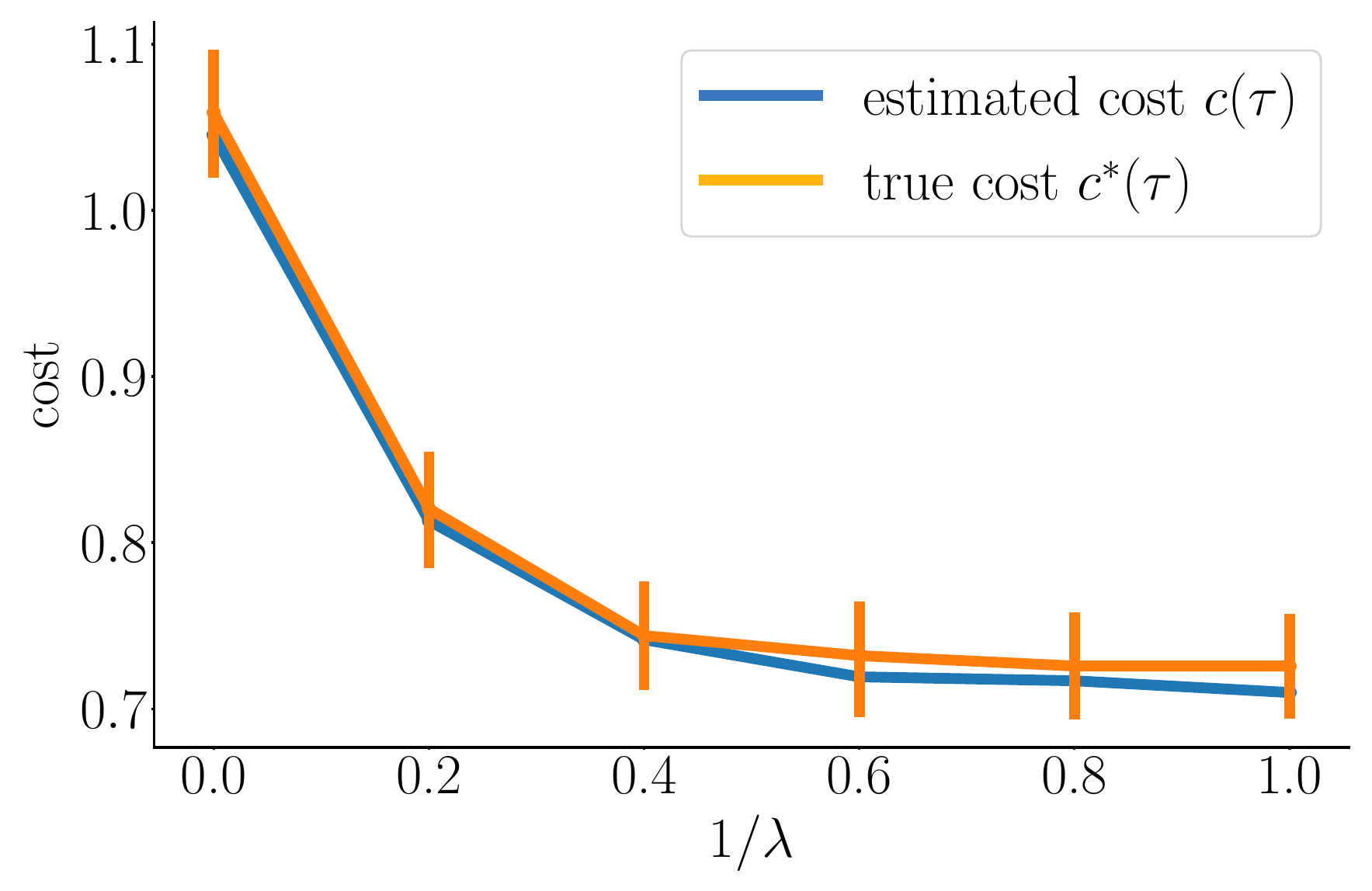}} \hspace{6mm}
 \subfloat[\% of misinfo vs $1/\alpha$]{\includegraphics[width = 0.25\textwidth]{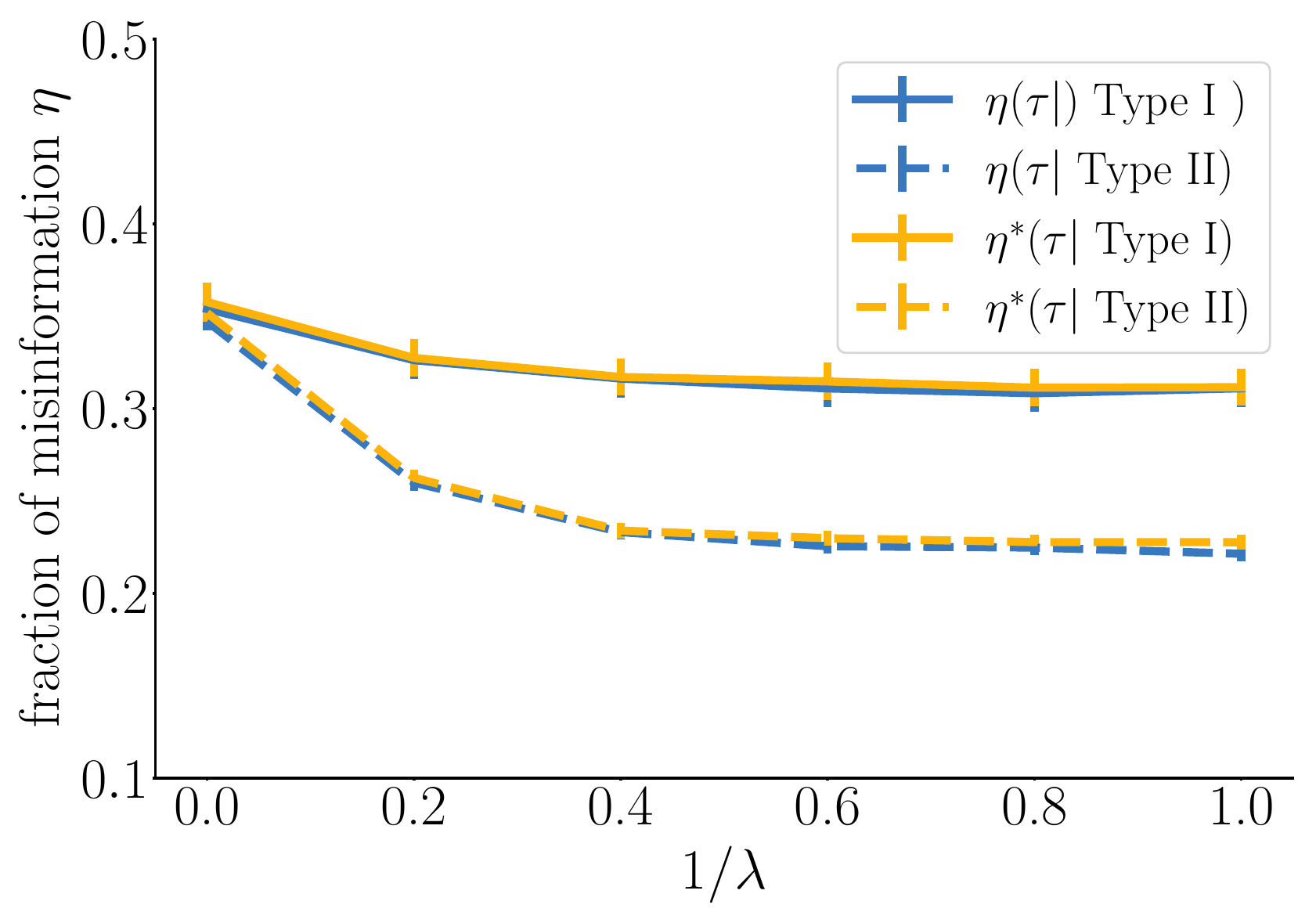}}
 \vspace{-1mm}
 \caption{Performance of the original ranking model ($1/\lambda \rightarrow 0$) and the optimal consequential ranking model ($1/\lambda > 0$) in synthetic data.
 Panels (a) and (b) show that, as $1/\lambda$ increases, the consequential ranking model is able to reduce the (true) cost to welfare without significantly decreasing
 the immediate utility.
 Panel (c) shows that, as $1/\lambda$ increases, the fraction of misinformation for viral posts on the top $3$ positions is lower than the fraction of misinformation
 for non viral posts.}\label{fig:syn_main}
 \vspace{-3mm}
\end{figure*}

\xhdr{Quality of the rankings}
We first compare the original ranking model $p_{\theta_0}$ and the optimal consequential ranking model $p^{*}_{\theta}$ in terms of three
quality metrics:
(i) the immediate utility $u(t)$; % , given by Eq.~\ref{eq:synthetic-u};
(ii) the cost to welfare $c(\tau)$; %  given by Eq.~\ref{eq:synthetic-c}; 
and, (iii) the true cost to welfare $c^{*}(\tau)$, defined as
%
% \begin{equation} \label{eq:synthetic-c-star}
$c^{*}(\tau) = \frac{1}{T} \sum_{t=1}^{T} \sum_{k=1}^{K=3} m_{\omega_k(t)}$.
% \end{equation}
%
Figure~\ref{fig:syn_main}(a-b) summarizes the results, where note that the original ranking model $p_{\theta_0}$ is just the optimal consequential ranking
model with $1/\lambda \rightarrow 0$.
The results show that:
(i) the consequential ranking model achieves lower (true) cost to welfare than the original ranking model; and,
(ii) as $1/\lambda$ increases, the consequential ranking model is able to reduce the (true) cost to welfare without significantly decreasing the immediate utility.
Next, we investigate whether the optimal consequential ranking model treats viral and non-viral posts differently.
Intuitively, the ranking model should be more willing to change the rank of high risk viral posts than that of high risk
non viral posts.
To confirm this intuition, we compute the fraction of estimated and true misinformation, $\eta(\tau)$ and $\eta^{*}(\tau)$, in the top $K = 3$ positions
of the rankings over time for both viral ($\alpha = 10$) and non viral ($\alpha = 0.1)$ posts. % , \ie,
%
% \manuel{saving space}
% \begin{equation*}
% \eta(\tau | \alpha) = \frac{ \sum_{t=1}^{T} \sum_{k=1}^{K=3} p_{\omega_k}(t) \II(\alpha_{\omega_k(t)} = \alpha) } { \sum_{t=1}^{T} \sum_{k=1}^{K=3} \II(\alpha_{\omega_k(t)} = \alpha) }
% \end{equation*}
% and
% \begin{equation*}
% \eta^{*}(\tau | \alpha) = \frac{ \sum_{t=1}^{T} \sum_{k=1}^{K=3} m_{\omega_k}(t) \II(\alpha_{\omega_k(t)} = \alpha) } { \sum_{t=1}^{T} \sum_{k=1}^{K=3} \II(\alpha_{\omega_k(t)} = \alpha) }.
% \end{equation*}
%
Figure~\ref{fig:syn_main}(c) summarizes the results, which show that, as we increase $1/\lambda$, the fraction of misinformation for viral posts on the top $3$ positions is lower
than the fraction of misinformation for non viral posts.
Appendix~\ref{app:synthetic} provides an in-depth comparison between optimal and P-L consequential ranking models.

%% file: 060real.tex
In this section, we compare the performance of ranking models that maximize an immediate measure of utility and parameterized consequential rankings
models using data from Reddit, a popular social news aggregation platform\footnote{\scriptsize Due to the size of the dataset, we were unable to run the 
weighted sampling procedure needed to implement optimal consequential rankings models.}.
%
% \manuel{saving space} , learned using Algorithm~\ref{alg:consequential-ranking},
%
Before we proceed further, we would like to acknowledge that: 
\begin{itemize}[noitemsep,nolistsep,leftmargin=0.8cm]
\item[(i)] Since we do not have access to the ranking algorithm used by Reddit (or any other social media platform), our experiments are a proof of concept, which 
demonstrate the practical potential of our methodology on real data using a simple P-L ranking model.
Evaluating the efficacy of our methodology across a wide range of deployed ranking algorithms is left as future work.

\item[(ii)] Our experiments use observational data, \ie, they are open loop. As a result, the rankings only influence the immediate utility and the cost of welfare 
but not the user dynamics.
%
% The user dynamics can only be changed if we make interventions in the system instead of relying on observational data since the users may post fewer contents that incure a cost to welfare if these contents
% are penalized in the ranking.
%
However, our evaluation is likely to be conservative---consequential rankings may achieve a greater reduction of the cost to welfare in an interventional experiment.
For example, in the context of uncivil behavior, there is empirical evidence that users are more likely to post uncivil comments if they are exposed to uncivil
comments before~\cite{cheng2017anyone, muddiman2017news}. Therefore, penalizing the rank of uncivil comments over time %, \ie, reducing their exposure, may
may prevent other users from engaging into uncivil behavior.
\end{itemize}

% Note that in Cheng et al., 2018, they write in the abstract that: Through an experiment simulating an online discussion, we find that both negative mood and seeing troll posts by others
% significantly increases the probability of a user trolling, and together double this probability.
% Also in Muddiman et al., they claim that news users engage with uncivil behavior
%

\xhdr{Dataset description and experimental setup}
We used a publicly available Reddit dataset\footnote{\href{https://archive.org/details/2015_reddit_comments_corpus}{https://archive.org/details/2015\_reddit\_comments\_corpus}.},
which contains (nearly) all publicly available comments to link submissions posted by Reddit users from October 2007 to May 2015.
In our experiments, we focused on the links submissions to the subreddit Politics and selected the set of submissions with more than $10$ and less than $60$ comments. After these 
preprocessing steps, our dataset comprised $3{,}173$ submissions and $68{,}016$ comments.

In a first set of experiments, we focus on the civility of the comments in each submission, as measured by an uncivility score $\phi$. In a second set of experiments, we focus on the misinformation 
spread by the comments of each submission, as measured by an unreliability score $\gamma$. Appendix~\ref{app:details-real-data} contains more details on the definition and estimation of 
both scores.
In both sets of experiments, we use $42{,}267$ comments from $1{,}973$ submissions as training set for learning the parameterized consequential ranking models
and $25{,}749$ comments from $1{,}200$ submissions as test set for evaluation.

% \xhdr{Experimental setup}
%
Each submission corresponds to one trajectory whose length $T$ is just the number of comments in the submission, \ie,
each time step corresponds to the time at which a new comment was created.
Then, at each time step $t \in \{0, \ldots, T\}$, the ranking model ranks the latest set of $n=10$ comments $\Ical(t)$.
Moreover, each comment $i$ has three features $\Xb_i(t) = [\tau_i, \phi_i, \gamma_i]$, where $\tau_i$ is the time (in seconds) elapsed since the
first comment was posted,
$\phi_i$ is the uncivility score and $\gamma_i$ is the unreliability score 
%
%the absolute value of the polarity of the comment if the polarity is negative and the mood of the comment is indicative or imperative
% and zero otherwise,
%
% and $\gamma_i$ is the absolute value of the unreliability score of the comment if the unreliability score is negative and zero otherwise.
%
%
% Appendix~\ref{app:details-real-data} provides a few examples of sentences with a high value of $\phi_i$, which measures uncivility, and $\gamma_i$, 
% which measures misinformation.
%
At each time $t$, the original ranking model $p_{\theta_0}$ aims to promote the most recent comment to the top of the ranking, \ie, its
immediate utility $u(t)$ is defined as 
%
% \begin{equation} \label{eq:synthetic-u-real}
$u(t) = \tau_{\omega_1(t)}$,
% \end{equation}
%
where $\omega_1(t)$ is the item at the top of the rank $\yb(t)$. To this aim, it uses a Plackett-Luce (P-L) model, given by Eq.~\ref{eq:pl}, with
$\theta = [1.25 \cdot 10^{-4}, 0, 0]$.
For the first set of experiments, the consequential ranking models $p_{\theta}$ aim to trade off fidelity to the original ranking model and the civility
of the comments on the top $K = 6$ positions of the rankings, \ie, 
%
% \begin{equation} \label{eq:real-uncivil-c}
$c(\tau) = \frac{1}{T} \sum_{t=1}^{T} \sum_{k=1}^{K=6} \phi_{\omega_k(t)}$.
% \end{equation}
%
In a second set of experiments, the consequential ranking models $p_{\theta}$ aim to trade off fidelity to the original ranking model and the misinformation
in the comments on the top $K = 6$ positions of the rankings, \ie, 
%
% \begin{equation} \label{eq:real-uncivil-c}
$c(\tau) = \frac{1}{T} \sum_{t=1}^{T} \sum_{k=1}^{K=6} \gamma_{\omega_k(t)}$.
% \end{equation}
%
In both cases, the consequential ranking models are Plackett-Luce (P-L) models, which we learned using Algorithm~\ref{alg:consequential-ranking}
with $M = 20$ iterations and $B = 100$ as batch size, and experiment with $\lambda=[0.0, 1.0, 2.0, ..., 10.0]$.

\begin{figure}[t]
\setlength{\tabcolsep}{0.05em}
 \centering
\subfloat[Uncivility]{ \includegraphics[width = 0.3\textwidth]{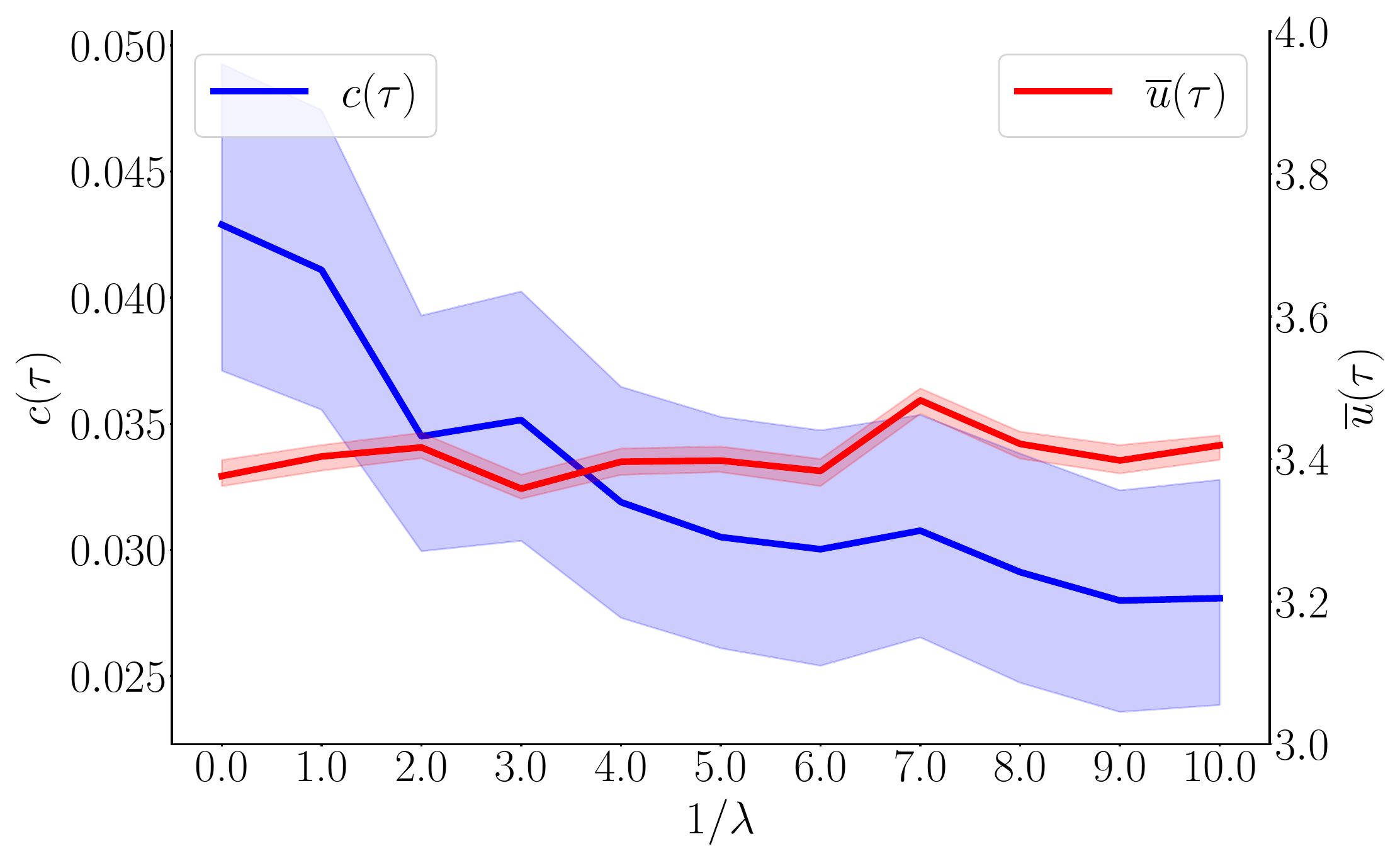} } \hspace{5mm}
\subfloat[Misinformation]{ \includegraphics[width = 0.3\textwidth]{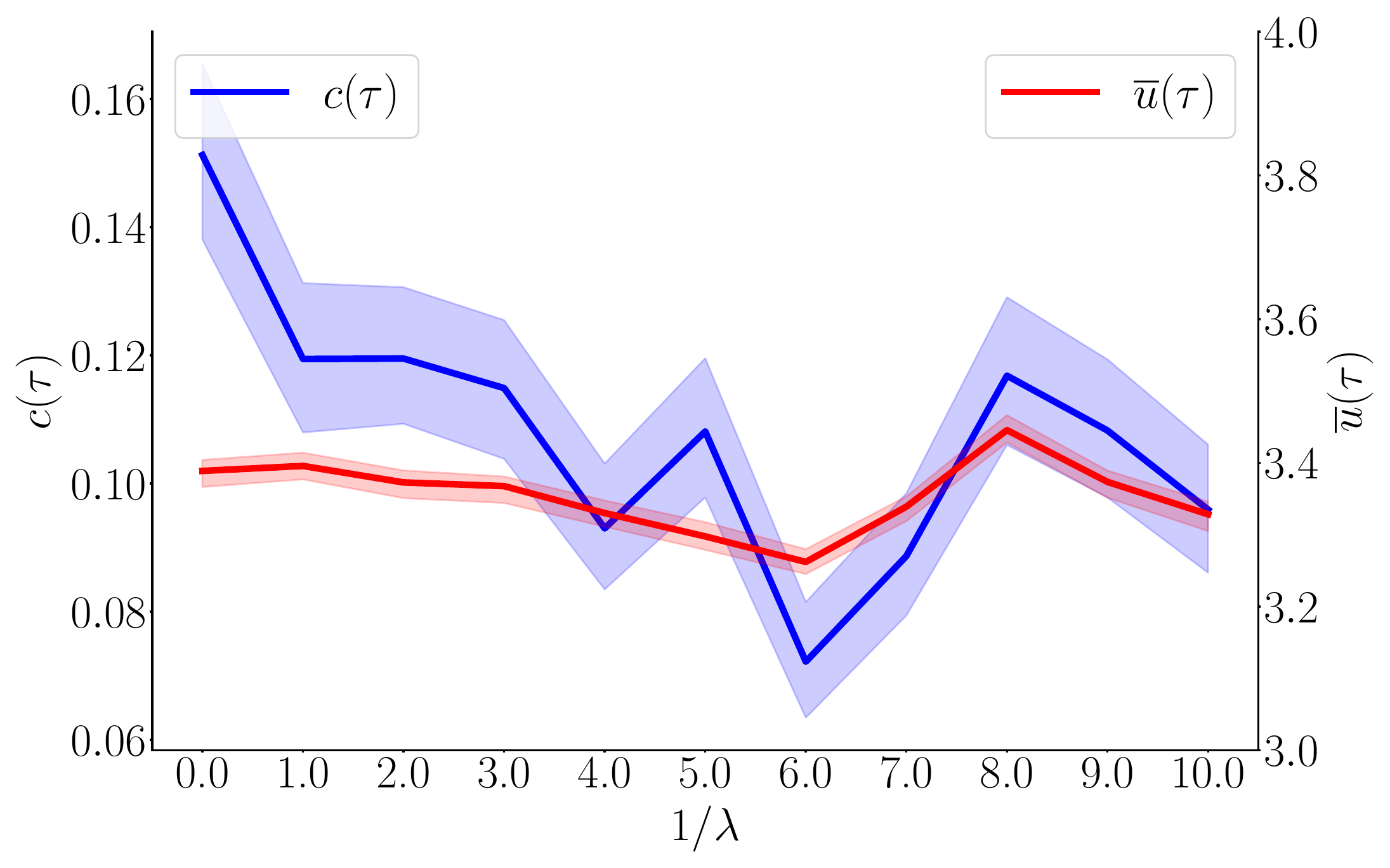} }
\vspace{-1mm}
 \caption{Performance of the original ranking model ($1/\lambda \rightarrow 0$) and the optimal consequential ranking model ($1/\lambda > 0$) in real
 Reddit data in terms of average immediate utility $\overline{u}(\tau)$ and cost to the welfare $c(\tau)$.
 %
 %In Panel (a), the cost to welfare measures the degree of uncivility of the top $K = 6$ ranking positions. In Panel (b), the cost to welfare measures the
 % amount of misinformation in the top $K = 6$ ranking positions. 
 Lines are averages and shaded areas are 95\% confidence intervals over all submissions in the test set.} \label{fig:real_reddit}
 \vspace{-4mm}
\end{figure}

\xhdr{Quality of the rankings}
We first compare the original ranking model $p_{\theta_0}$ and the consequential P-L ranking models $p_{\theta}$ using the average immediate utility and
the cost to welfare. Here, note that, in the first set of experiments, the cost to welfare measures the degree of uncivility of the top ranking
positions while, in the second set of experiments, it measures the amount of misinformation.
Figure~\ref{fig:real_reddit} summarizes the results,
where note that the original ranking model $p_{\theta_0}$ is just the optimal consequential ranking model with $1/\lambda \rightarrow 0$.
The results show that the consequential P-L ranking models are able to reduce the cost to the welfare up to $30$\% at a minimum cost in terms of immediate
utility---they are able to reduce the degree of uncivility and the amount of misinformation at the top ranking positions without significant changes to the original
reverse chronological ranking.

%% file: 070conclusions.tex
We have initiated the design of (parameterized) consequential ranking models that optimally trade off between
the fidelity to ranking models optimizing for immediate utility and the long-term welfare.
%
% More specifically, we have first introduced a joint representation of rankings and user dynamics using Markov
% decisions processes.
%
% Exploiting this representation, we have shown that we can obtain optimal consequential rankings just by applying
% weighted sampling on the rankings provided by the model optimizing for immediate utility. However, in practice, such
% a strategy may be inefficient and impractical, specially in high dimensional scenarios.
%
% To overcome this, we introduced an efficient gradient-based algorithm to learn parameterized consequential ranking
% models that effectively approximate the optimal ones.
%
% Finally, we have experimented on synthetic and real data to show the efficacy of our parameterized consequential
% ranking models.
%
Our work opens up many interesting avenues for future work.
For example, we have considered probabilistic ranking models and a fidelity measure based on KL divergence. A natural next 
step is to augment our methodology to allow for deterministic ranking models and consider other fidelity measures between 
rankings. 
%
% For simplicity, among the many families of parameterized (probabilistic) ranking models, we have chosen a simple Plackett-Luce
% (P-L) ranking model. 
It would be very interesting to apply our framework to more sophisticated ranking models.
Moreover, we have assumed that the models that optimize for immediate utility are optimal.
% that our models that optimize
However, they may be suboptimal in terms of the sum of immediate utility over time since it is unclear that current ranking models 
are designed to account for the consequences that their proposed rankings have on the feature matrices. It would be very interesting 
to account for this in future work.
%
% In our experiments, at each time step, each item has an independent (and additive) cost to the welfare. 
It would also be interesting to experiment with settings in which the cost to welfare cannot be factorized into items, \eg, information diversity.
Finally, we have evaluated our algorithm using observational real data, however, it would be very revealing to perform an evaluation
based on interventional experiments.

%% file: 080appendix.tex
%\section{Our joint representation of rankings and user dynamics} \label{app:model}
%
\section{Graphical representation of rankings and user dynamics} \label{app:model}
\begin{figure*}[!h]
 \centering
 \includegraphics[width=.7\textwidth]{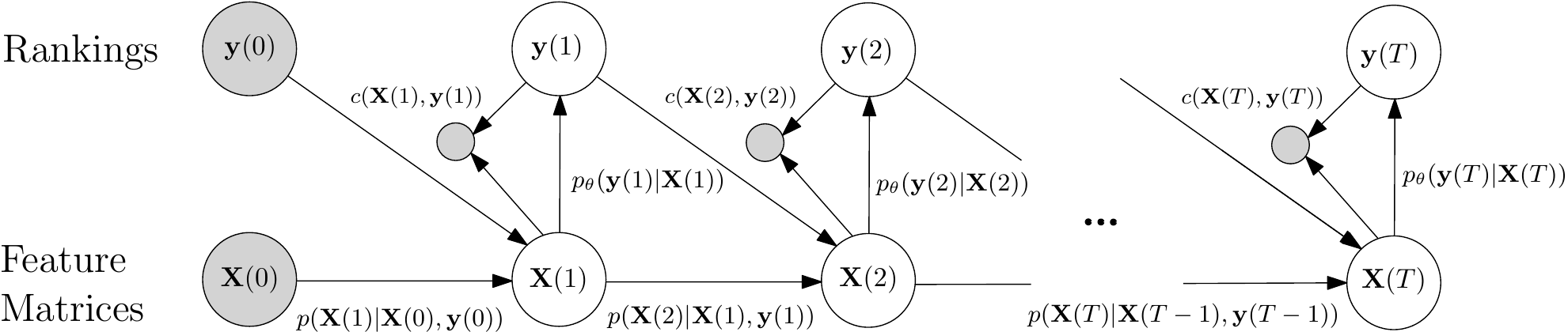}
 \caption{Our joint representation of rankings and user dynamics using Markov decision processes (MDPs).
 The ranking model $p_{\theta}(\yb(t) \given \Xb(t))$ provides a ranking $\yb(t)$ for a set of items on the basis of the feature matrix $\Xb(t)$ of the items and
 both the feature matrix and the provided ranking result in a cost to welfare $c(\Xb(t), \yb(t))$.
 The distribution of user dynamics $p(\Xb(t+1) \given \Xb(t), \yb(t))$ determines the feature matrix $\Xb(t+1)$ on the basis of the previous feature matrix
 $\Xb(t)$ and ranking $\yb(t)$.} \label{fig:representation}
 \vspace{-3mm}
\end{figure*}

\section{Algorithm for sampling from optimal consequential ranking} \label{app:sampling-alg}
Algorithm~\ref{alg:consequential-ranking-sampling} shows the stratified sampling~\cite{douc2005comparison} method for obtaining samples from optimal consequential ranking. Within the algorithm, \textsc{Sample}$(p_{\theta_0}, \kappa)$ samples $\kappa$ trajectories from $p_{\theta_0}(\tau)$ and
\textsc{StratifiedSampler}$(\Dcal', W)$ generates $|\Dcal'|$ samples weighted by $W$ using stratified sampling.

\begin{algorithm}[H]
\small
\caption{It samples from an optimal consequential ranking model given $p_{\theta_0}$.}
\label{alg:consequential-ranking-sampling}
  \begin{algorithmic}[1]
    \Require Cost to welfare $c(\cdot)$, parameter $\lambda$, original ranking model $p_{\theta_0}$, $(\Xb(0), \yb(0))$, \# of samples $B$, \# of samples $\kappa$ to compute $G[Z_{T}]$.
   \Statex
      \State $\Dcal \gets \Call{Sample}{{p_{\theta_0}, \kappa}}$ \Comment{samples for estimating $G[Z_{T}]$.}
      \State $G[Z_{T}] \gets 0$
      \For{$c(\tau_i) \in \Dcal$}
              \State $G[Z_{T}] \gets G[Z_{T}] + \exp{\left(-\lambda^{-1} c(\tau_i)\right)}/\kappa$
      \EndFor
      \State $\Dcal' \gets \Call{Sample}{{p_{\theta_0}, B}}$ \Comment{unweighted samples.}
      \State $W \gets []$ \Comment{array of weights}.
      \For{$c(\tau_i) \in \Dcal'$}
              \State $W[i] \gets \exp{\left(-\lambda^{-1} c(\tau_i)\right)}/\kappa/G[Z_{T}]$
      \EndFor
      \State $W \gets W/\Call{Sum}{W}$
      \State \Return $\Call{StratifiedSampler}{{\Dcal', W}}$
  \end{algorithmic}
\end{algorithm}

\section{Additional experiments on synthetic data} \label{app:synthetic}
\begin{figure*}[h]
\setlength{\tabcolsep}{0.05em}
 \centering
 \subfloat[Immediate utility vs time]{\includegraphics[width = 0.28\textwidth]{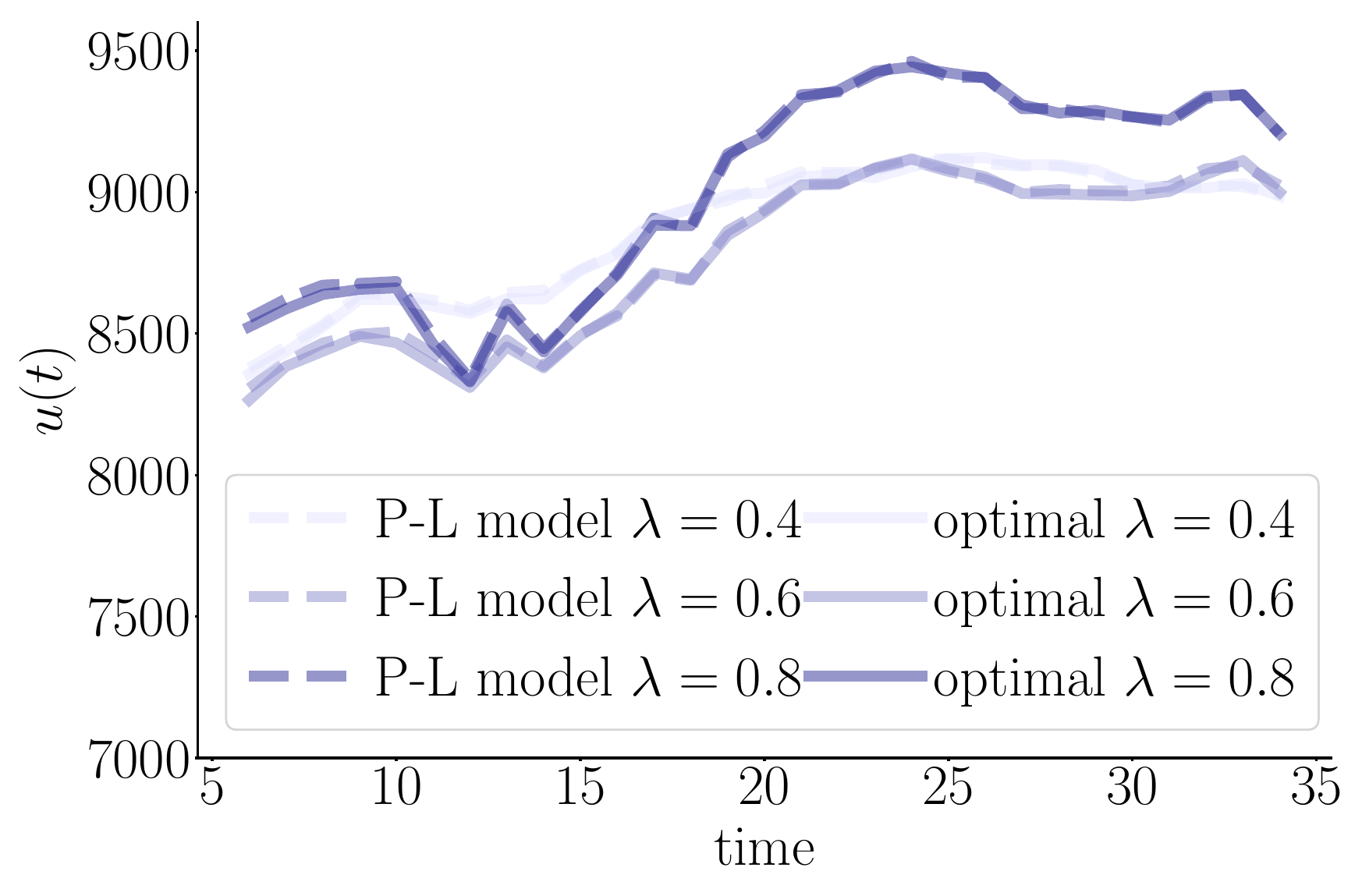}} \hspace{4mm}
 \subfloat[Cost to welfare vs $1/\lambda$]{\includegraphics[width = 0.28\textwidth]{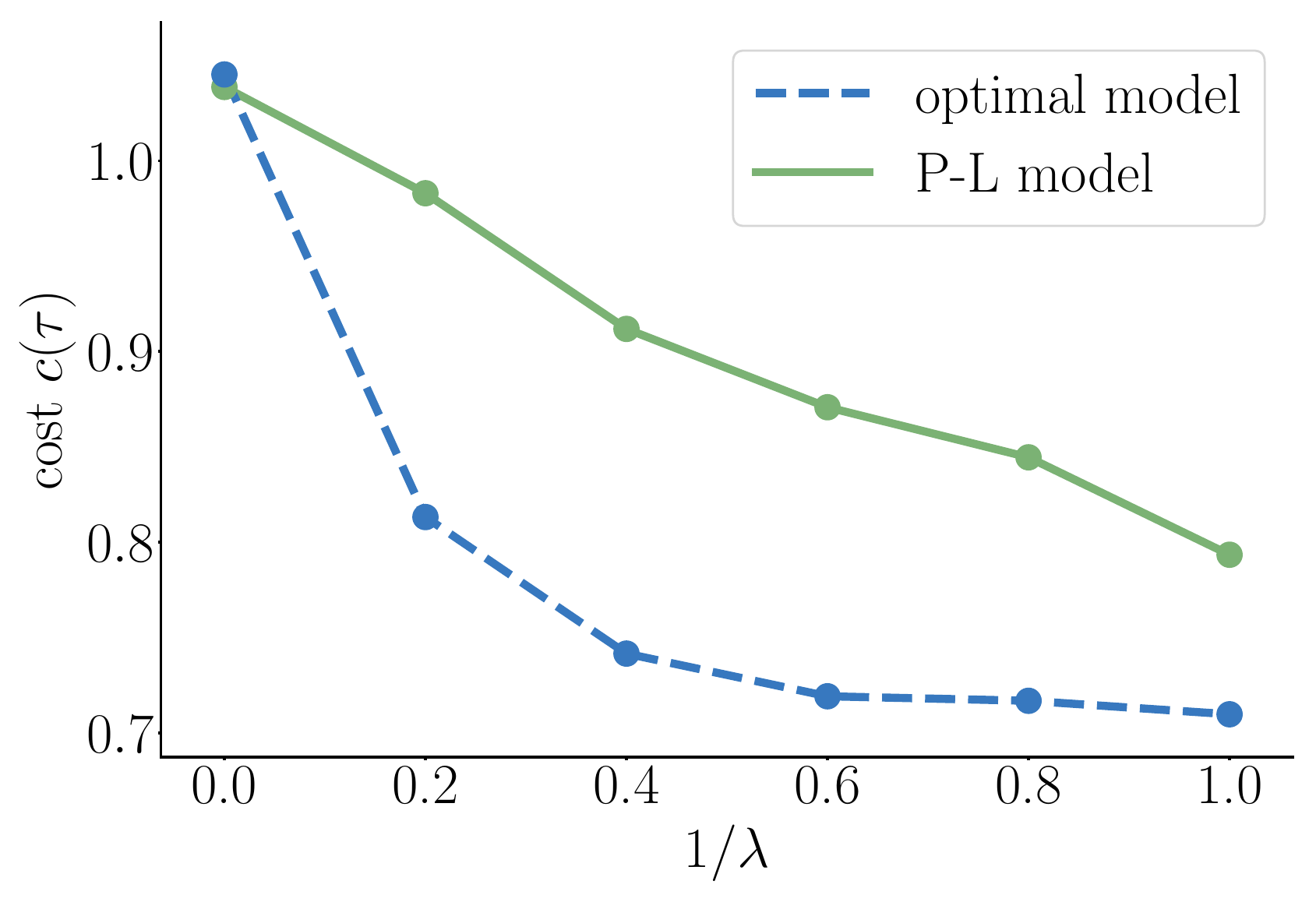}} \hspace{4mm}
 \subfloat[Average \# samples per ranking]{\includegraphics[width = 0.28\textwidth]{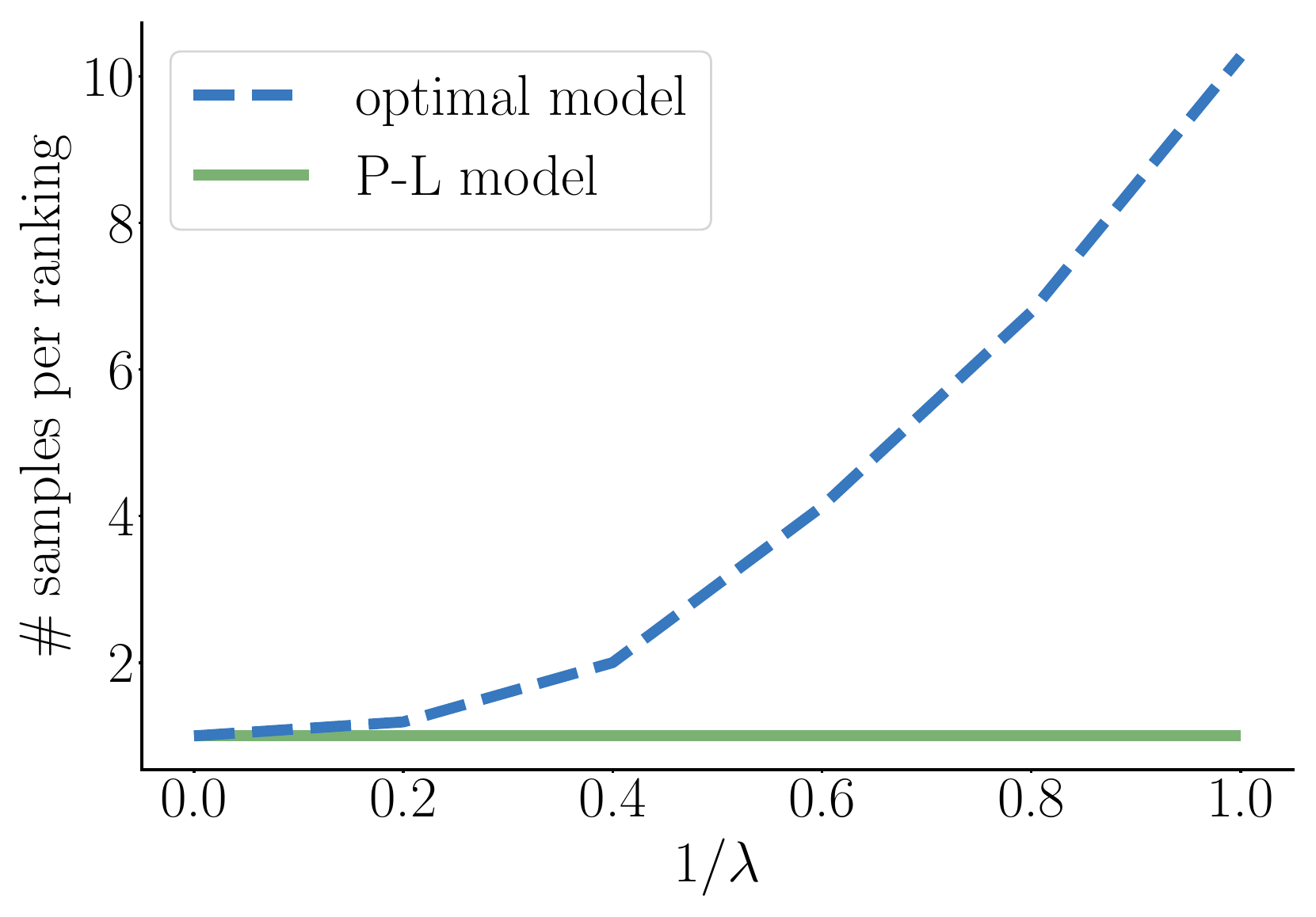}}
 \caption{Performance and scalability of the optimal consequential ranking model and the P-L consequential ranking model learned using Algorithm~\ref{alg:consequential-ranking}.
Panels (a-b) show that the optimal consequential ranking model achieves a better trade off between immediate utility and cost to the welfare than
the P-L consequential ranking model. However, Panel (c) shows that the weighted sampling the optimal model uses to generate rankings quickly
become computationally prohibitive in terms of \# samples needed per ranking as $1/\lambda$ increases and the difference between the original
ranking model and the optimal consequential ranking model increases.
}\label{fig:rejection_vs_reinforce}
\end{figure*}

First, we compare the performance of the optimal consequential ranking model computed via weighted sampling and the P-L ranking model learned
using Algorithm~\ref{alg:consequential-ranking} using the same quality metrics as in the previous section.
Figure~\ref{fig:rejection_vs_reinforce}(a-b) summarizes the results.
We observe that both the optimal and P-L consequential ranking models achieve similar values of immediate utility over time. However,
the optimal model has a competitive advantage in terms of cost to the welfare, which becomes more pointed as $1/\lambda$ grows.
These findings suggest that, the larger the value of $1/\lambda$, the more difficult is to learn a P-L model that approximates
effectively the optimal model.

Next, we compare the scalability of both models in terms of the number of samples needed per ranking. Figure~\ref{fig:rejection_vs_reinforce}(c)
summarizes the results, which shows that, as $1/\lambda$ grows, it becomes computationally prohibitive to generate optimal consequential rankings
using weighted sampling due to the growing difference between $p^{*}_{\theta}$ and $p_{\theta_0}$.
This questions the practicality of weighted sampling to generate optimal consequential rankings.

\section{Uncivility and unreliability scores} \label{app:details-real-data}
To estimate the uncivility score $\phi$ for each comment, we apply sentiment analysis on the text of the comments using the software package \emph{Pattern}\footnote{\href{https://www.clips.uantwerpen.be/pages/pattern-en}{https://www.clips.uantwerpen.be/pages/pattern-en}} and, for each comment, obtain two quantities: mood and polarity.
The mood of a comment can take one of the following four values: indicative, imperative, conditional and subjunctive. The polarity of a comment is a real number in $[-1,1]$, where
lower (higher) values indicate more negative (positive) words in the text.
Then, we define the uncivility score $\phi$ of a comment as the absolute value of the polarity of the comment if the polarity is negative and the mood of the comment is indicative 
or imperative and zero otherwise

We estimate the unreliability score $\gamma$ for each comment by estimating the average unreliability score of the domains that appeared in each of them, as estimated by aggregating 
publicly available data from Politifact and Snopes\footnote{\href{https://www.kaggle.com/arminehn/rumor-citation/version/3}{https://www.kaggle.com/arminehn/rumor-citation/version/3}}.
More specifically, our combined dataset contains fact checking information for $17{,}804$ unique urls from $4{,}540$ unique domains.
For each url, it assigns a label that indicates the reliability of its content. We used these labels to assign a numerical unreliability score for each url.
More specifically, if the url is labeled as ``false'', ``pants-fire'', ``mfalse'' or ``legend'', we set the unreliability score to $1$.
If the url is labeled as ``true'', ``mtrue'' or ``mostly-true'', then we set the unreliability score to $-1$.
And, if the url is labeled using some other label value, we set the unreliability score to $0$.
We computed an unreliability score for each domain, which measures its level of (un)trustworthiness, by taking the average of the unreliability scores of individual
urls from the domain.
Then, we define the unreliability score $\gamma$ of a comment as the average unreliability score of the domain(s) of the link(s) used in the comment if the average
is negative and zero otherwise.
Here, also note that, if a comment does not contain any links or the domain(s) of the link(s) does not appear in our dataset, we set the unreliability score for that
comment to $0$.

Tables~\ref{tbl:example-uncivility} and~\ref{tbl:example-misinformation} provide a few examples of comments with a high uncivility score and domains with a high 
unreliability score.
\begin{table}[h]
\small
\centering
\ra{1.3}
\begin{tabular}{@{}lr@{}}\toprule
Comment & Uncivility ($\phi$) \\
\midrule
If you once tell a lie, the truth is ever after & 0.0\\[-1.2ex]
 your enemy. & \\
I dream of a world where your bigoted stupid & 0.1\\[-1.2ex]
ideas don't have the popular shield of faith. & \\
Shut the f**k up and die already you POS  &  0.4 \\[-1.2ex]
 warmongering profiteer. & \\
Crap? Or pap. Take your pick. & 0.8 \\
i blame the evil KOCH BROTHERS! & 1.0 \\
\bottomrule
\end{tabular}
\caption{Examples of sentences with different levels of uncivility, as estimated by the feature $\phi$. Comments with higher levels of uncivility typically correspond to those that
use foul language. \label{tbl:example-uncivility}}
\end{table}
\begin{table}[h]
\small
\centering
%\ra{1.3}
\vspace{-6mm}
\begin{tabular*}{0.56\textwidth}{l@{\hskip 1.3in}r}\toprule
Url & Misinformation ($\gamma$) \\ \midrule
aids.gov  &\  0.0 \\
pbs.org & 0.26 \\
breitbart.com & 0.56 \\
lifeisajoke.com	 & 1.0\\
\bottomrule
\end{tabular*}
\caption{Examples of domains that spread different amounts of misinformation, as estimated by the feature $\gamma$. \label{tbl:example-misinformation}}
\end{table}

\section{Additional experiments on real data} \label{app:experiments-real-data}
\begin{figure}[h]
\setlength{\tabcolsep}{0.05em}
 \centering
  \subfloat[Uncivility]{\includegraphics[width = 0.36\textwidth]{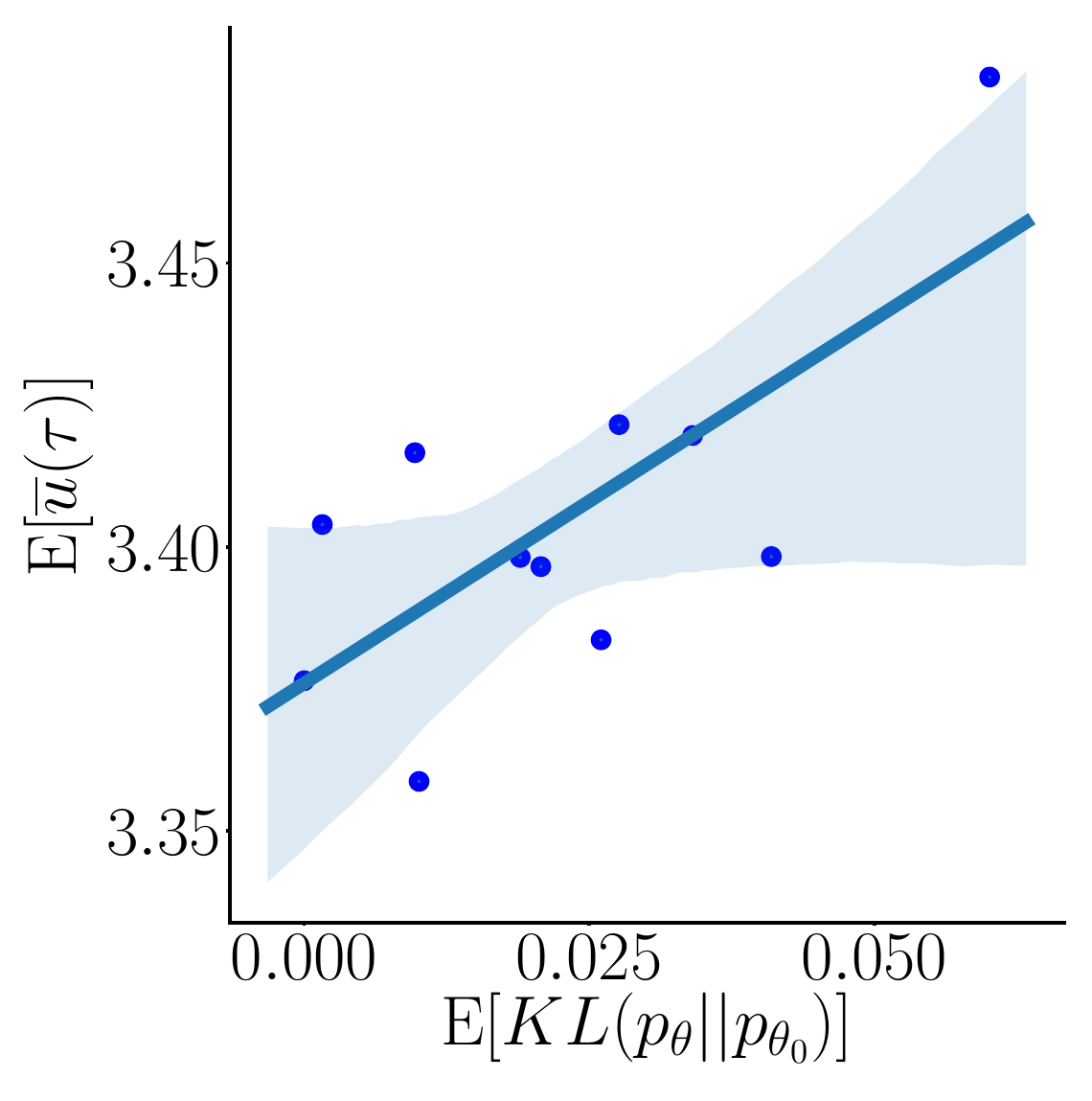}} \hspace{4mm}
  \subfloat[Misinformation]{\includegraphics[width = 0.36\textwidth]{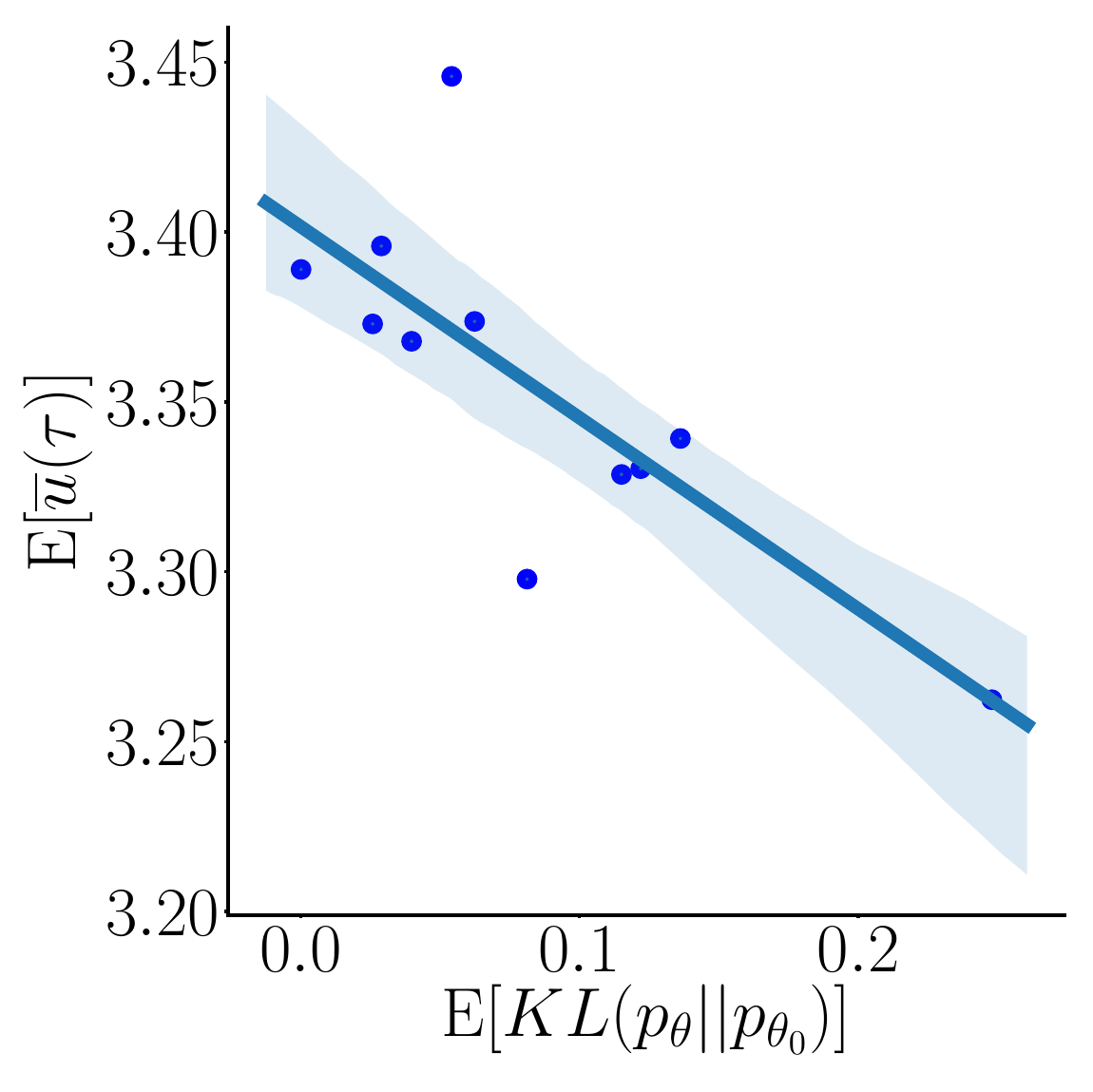}}
   \vspace{-2mm}
 \caption{Average immediate utility $\EE[\overline{u}(\tau)]$ achieved by the consequential ranking model vs the average Kullback-Leibler (KL) divergence
 $\EE[KL(p_{\theta} \,||\, p_{\theta_0} )]$ between its induced probability $p_{\theta}(\tau)$ and the  probability $p_{\theta_0}(\tau)$ induced by the original
 ranking model.} \label{fig:real_loss_kl_div_reddit}
 \vspace{-3mm}
\end{figure}

In the experiments on real data in the main paper, we have assumed that the original ranking model $p_{\theta_0}$ maximizes immediate utility. Moreover, we have taken for granted 
that there is a negative correlation between the immediate utility achieved by the consequential ranking model and the Kullback-Leibler (KL) divergence between its induced probability 
$p_{\theta}(\tau)$ and the probability $p_{\theta_0}(\tau)$ induced by the original ranking model.
However, in practice, it may happen that the original ranking model is not optimal in terms of immediate utility maximization. This may be specially the case if the original
ranking model greedily maximizes immediate utility and does not take into account how their proposed rankings change the user dynamics.
Figure~\ref{fig:real_loss_kl_div_reddit} demonstrates that, in our first set of experiments, there is indeed a negative correlation between the immediate utility and the KL
divergence, however, in our second set of experiments, there is a positive correlation. This suggests that the original ranking model in the second set of experiments is
suboptimal.
